\documentclass[sigconf]{acmart}
\AtBeginDocument{%
  }

\setcopyright{acmlicensed}
\copyrightyear{2018}
\acmYear{2018}
\acmConference{Human-centered Explainable AI Workshop (HCXAI) @ CHI 2026} 

\begin{document}

\title{Not All Explanations Are Sought: Information-Seeking Psychology for Human-Centered XAI}

\author{Andrea Beretta}
\email{andrea.beretta@isti.cnr.it}
\orcid{0000-0001-8531-9325}
\author{Salvatore Rinzivillo}
\email{rinzivillo@isti.cnr.it}
\orcid{0000-0003-4404-4147}
\affiliation{%
  \institution{CNR-ISTI}
  \city{Pisa}
  \state{}
  \country{Italy}
}

\renewcommand{\shortauthors}{Beretta et al.}

\begin{abstract}
This position paper argues that human-centered explainable AI (HCXAI) should incorporate insights from the psychology of information seeking. Drawing on Sharot and Sunstein's framework of information-seeking motives, we propose that people evaluate whether to engage with explanations based on three types of expected utility: instrumental (will it help me act better?), hedonic (will it make me feel better?), and cognitive (will it improve my understanding?). Each utility is estimated through a lens shaped by well-documented cognitive biases, including illusion of control, automation bias, unrealistic optimism, impact bias, overconfidence, and confirmation bias. These biases can lead to two failure modes: excessive information-seeking that fragments attention without improving decisions, and insufficient information-seeking that leaves critical risks and misunderstandings unexamined. This challenge is particularly acute for agentic AI systems, where explanations must support not just understanding a single output but anticipating cascading actions, assessing risks, and deciding when to intervene. By integrating information-seeking psychology into HCXAI, we advocate for a shift from making explanations available to making them sought: designing systems that account for when and why users actually want to know.
\end{abstract}




\maketitle

\section{INTRODUCTION}
The field of Human-Centered Explainable AI (HCXAI) has made significant progress in understanding what to explain and how to present explanations to users. Nevertheless, a fundamental question remains underexplored: why and when do people actually seek explanations from AI systems?

Current XAI research implicitly assumes that users want explanations and will engage with them when provided \cite{ArrietaRSBTBGGM20}, supporting the idea that "knowledge is always valuable" \cite{grant1998intrinsic}. However, recent research showed a different picture, and there has been growing interest in those who use the explanation \cite{ehsan2024xai} and in whether the explanations are always valuable \cite{colaner2022explainable}. 

Emerging evidence from the HCXAI community itself points to a more complex reality. LLM-generated natural language explanations, while more readable, can reinforce overreliance rather than foster critical engagement \cite{suh2025don}. 
Time pressure increases compliance with AI suggestions regardless of explanation quality. \cite{shahu_2025_15170490},
and users exhibit dramatically different engagement profiles from minimal to deeply exploratory  \cite{freiberger_2025_15170451} revealing that engagement with explanations varies dramatically across individuals and contexts. 

These findings converge on a shared but undertheorized insight: making explanations available does not ensure they will be sought, and making them more accessible does not guarantee they will be scrutinized. What is missing is a unifying theoretical framework that explains when and why users engage with, or avoid, AI explanations.

People are selective information seekers. They do not pursue all available knowledge; they seek or avoid information based on its expected impact on their actions, emotions, and understanding of the world \cite{case2005avoiding}. This selectivity is not a failure of attention or literacy, it is a fundamental feature of human cognition that XAI research has yet to systematically address.

This gap is particularly critical for agentic AI systems, where explanations are not just about understanding a single output, but about anticipating cascading actions, assessing risks, and deciding whether to intervene \cite{chaduvula2026features}. Recent work has begun to address these challenges from the explanation-design side, proposing causal and multi-level explanations that span strategic, tactical, and operational levels of AI agent activity \cite{ferrario_2025_15170379}, as well as explanations that must drive user action in safety-critical infrastructure \cite{folstad_2025_15170429}. 
However, even the most sophisticated explanation designs will fail to support oversight if users do not seek them. If users avoid explanations that induce negative affect, ignore information they believe will not change their actions, or dismiss explanations they consider redundant, even the most transparent AI system will fail to support appropriate oversight.

In this position paper, we argue that HCXAI should integrate insights from the psychology of information seeking. Drawing on Sharot and Sunstein's framework of information-seeking motives \cite{sharot2020people}, we propose that explanations are not inherently valuable, their perceived utility depends on instrumental, hedonic, and cognitive factors. We outline how well-documented cognitive biases distort users' evaluation of these utilities, producing predictable patterns of excessive or insufficient information seeking. This framework offers a theoretical lens for understanding phenomena that the HCXAI community has increasingly observed, from overreliance and explanation avoidance to heterogeneous engagement profiles, and provides a foundation for designing XAI systems that account for when and why users actually want to know.

\section{HOW PEOPLE DECIDE WHAT THEY WANT TO KNOW}
Understanding how people seek information is essential for designing explanations they will actually use. Sharot and Sunstein (2020) \cite{sharot2020people} propose an integrative framework in which information-seeking is driven by three distinct utilities:
\begin{itemize}
    \item \textbf{Instrumental utility}  reflects whether information will help the person make better decisions or take more effective actions. People seek information they believe will improve outcomes and avoid information they perceive as useless for action.
    \item \textbf{Hedonic utility} reflects the anticipated emotional impact of knowing. People tend to seek information when they expect good news and avoid it when they expect bad news  using information strategically to regulate their emotions.
    \item \textbf{Cognitive utility} reflects whether information will strengthen one's mental model of the world. People are drawn to information that connects to what they already care about, and that reduces meaningful uncertainty.
\end{itemize}

According to this framework, the estimated impact of information on each of these three dimensions can be \textbf{positive} (prompting information-seeking), \textbf{negative} (inducing active avoidance), or \textbf{zero} (leading to indifference).
Critically, people do not assess instrumental, hedonic, and cognitive utilities accurately, and their estimates are shaped by well-documented cognitive biases, which can distort information-seeking behavior in predictable ways.

Biases affecting instrumental utility include the \textit{illusion of control}, i.e. the tendency to overestimate one's ability to influence outcomes through action \cite{langer1975illusion}, and \textit{automation bias}, that is, the tendency to over-rely on automated systems and discount contradictory information \cite{mosier2018human,parasuraman2010complacency} reducing the perceived value of understanding why a system acts (or looking for more information about the system), since users trust it to perform correctly. 

Biases affecting hedonic utility include \textit{unrealistic optimism,} i.e., the tendency to overestimate the probability of positive outcomes, which reduce motivation to seek information about risks \cite{weinstein1980unrealistic}. The \textit{impact bias}, the tendency to overestimate the intensity and duration of future emotional reactions \cite{wilson2005affective}, can cause information avoidance about failures or uncertainties.

Biases affecting cognitive utility include \textit{the illusion of knowledge}, that is, the underestimating how much new information would change one's understanding \cite{kruger1999unskilled,sloman2017knowledge}, and \textit{confirmation bias}, or the tendency to seek information that validates existing beliefs while avoiding disconfirming evidence \cite{nickerson1998confirmation,sharma2024generative}. These biases lead users to dismiss explanations as redundant or to seek only for information that confirms what they already know and believe.
Table \ref{tab:Info_seek_utilities} summarizes how each utility, its associated biases, and their potential consequences for XAI design.



 \begin{table}[ht]
  \centering
  \caption{Information-seeking utilities, associated biases, and implications for XAI.}
  \label{tab:Info_seek_utilities}
  \scriptsize 
  \setlength{\tabcolsep}{2pt} 
  \renewcommand{\arraystretch}{1.5} 
  \begin{tabularx}{\columnwidth}{@{} >{\raggedright\arraybackslash}l 
                                   >{\raggedright\arraybackslash}X 
                                   >{\raggedright\arraybackslash}X 
                                   >{\raggedright\arraybackslash}X @{}}
    \toprule
    \textbf{Utility} & \textbf{Questions the user asks} & \textbf{Associated biases} & \textbf{Consequence for XAI} \\
    \midrule
    \textbf{Instrumental} & "Will this help me decide or act?" & Illusion of control, \newline Automation bias & Over-seeking (false confidence in intervention ability) or under-seeking (over-trust in the system) \\
    \addlinespace
    \textbf{Hedonic} & "How will knowing this make me feel?" & Unrealistic optimism, \newline Impact bias & Avoidance of explanations about risks, failures, or uncertainty \\
    \addlinespace
    \textbf{Cognitive} & "Does this fit what I already know?" & Illusion of knowledge, \newline Confirmation bias & Dismissing explanations as redundant; seeking only confirming information \\
    \bottomrule
  \end{tabularx}
\end{table}

\section{IMPLICATIONS FOR EXPLAINABLE AI}
Much of current XAI research operates under an implicit assumption: that providing information equates to providing knowledge, and that knowledge is inherently valuable. This assumption is reflected in design choices that prioritize completeness and transparency the more information a system discloses, the better. However, this "information is knowledge" view overlooks a critical insight from cognitive psychology: not all information is perceived as useful, and not all knowledge is sought. 

People do not passively engage with all available information; they actively decide what they want to know based on an evaluation of expected utilities \cite{sharot2020people}. 
This distinction has profound implications for how we design explainable AI systems.

Applying Sharot and Sunstein's framework of information-seeking to XAI reveals how incorrect way people use to look for information can distort users' evaluation of explanations, producing two failure modes:

\begin{itemize}
    \item \textbf{Excessive information-seeking} that fragments attention without improving decisions. E.g., the \textit{illusion of control} can lead the users to seek explanations under the assumption that they can intervene,
    \item \textbf{Insufficient information-seeking} that leaves critical risks and misunderstandings unexamined. 
    \end{itemize}

These failure modes can be attributed to the following cognitive dynamics:

\begin{itemize}
    \item \textbf{Seeking explanations that cannot improve decisions}. The \textit{illusion of control} leads users to request explanations under the assumption that understanding will enable intervention, even when no intervention is possible. This dynamic is reinforced by a subtle cognitive mechanism: explanations increase the subjective perception of understanding, which is then mistakenly equated with the capacity to intervene and ultimately driving users to overestimate the value of the information received. For example, a user interacting with an LLM-based forecasting agent might repeatedly ask why the system predicts a negative outcome for a given scenario, believing that understanding the model's reasoning will allow them to act and reverse that outcome. In reality, the outcome depends on external variables entirely beyond the user's control. The explanation feels empowering, but no intervention is available. Attention has been spent, explanations have been sought, but no decision has improved."
    \item \textbf{Avoiding explanations that reveal uncomfortable truths} \textit{Unrealistic optimism} and \textit{impact bias} lead users to avoid information about risks, failures, or uncertainties. A user interacting with an AI assistant for financial planning might skip explanations flagged as "limitations" or "assumptions" not because these are irrelevant, but because engaging with them would induce anxiety. The result is blind spots precisely where scrutiny is most needed.
     \item \textbf{Trusting without verifying} \textit{Automation bias} reduces the perceived need to understand why a system acts. If users believe the AI is generally reliable, they may see little value in reading explanations at all: "It usually gets things right, so why bother?" This leads to passive acceptance rather than informed oversight.
     \item \textbf{Seeking only what confirms}\textit{ Confirmation bias} drives users toward explanations that validate their existing expectations and away from information that challenges them. A user who believes an AI recommendation aligns with their intuition may engage deeply with supporting rationale while dismissing caveats or alternative interpretations.
     \item \textbf{Dismissing explanations as redundant} The \textit{illusion of knowledge} leads users to believe they already understand how a system works, when in fact their mental model is incomplete or inaccurate. A user who has interacted with an AI assistant for weeks may assume they "know how it thinks" and skip explanations entirely, "I already understand this." However, their intuitions may be based on superficial patterns rather than actual system behavior. When the system acts unexpectedly, they lack the conceptual foundation to diagnose what went wrong. The explanation was available, but never sought. 

\end{itemize}

Consider a concrete example in the context of agentic AI systems. A user interacts with an LLM-based agent to complete a complex task like drafting a document, planning a project, or analyzing data. The agent, designed to be transparent and helpful, provides extensive explanations at each step: why it chose a particular approach, what alternatives it considered, what assumptions it made. From a traditional XAI perspective, this is exemplary behavior. Yet the user may find themselves overwhelmed, unable to complete the task because they are continuously processing information that does not serve their immediate goals. The explanations have high transparency but low perceived instrumental utility and they do not help the user act more effectively. Worse, the cognitive load induced by excessive information may reduce hedonic utility (generating frustration) and fragment the user's cognitive utility (disrupting rather than strengthening their mental model of the task). The result is a paradox: more explanation leads to worse outcomes. This is not a failure of the explanation's content, but a failure to consider whether and when the user needs that information.

The broader implication concerns the very possibility of effective human-AI teaming. Successful collaboration between humans and AI systems particularly agentic systems that plan, act, and adapt requires a shared understanding of goals, actions, and constraints. Explanations are meant to support this shared understanding. However, if explanations are designed without considering how users evaluate their utility, a fundamental mismatch emerges: the system provides information that users do not seek, while failing to address the information they actually need. Users may disengage from explanations entirely, leading to blind over-reliance; or they may seek excessive clarification, fragmenting their attention and disrupting task completion. In neither case does genuine teaming occur. To achieve appropriate human-AI collaboration, XAI must move beyond the assumption that transparency is intrinsically valuable. Instead, it must consider the perceived instrumental, hedonic, and cognitive utility of explanations designing not just for what users should know, but for what users will actually want to know, and why.

\section{FUTURE DIRECTIONS}

In this paper, we have argued that human-centered XAI should move beyond the implicit assumption that explanations are inherently valuable. Drawing on the psychology of information seeking, we proposed that users evaluate explanations based on their perceived instrumental, hedonic, and cognitive utility and that predictable biases distort these evaluations in ways that can undermine both appropriate reliance and effective human-AI teaming.

We offer a foundational reframing that shifts the unit of analysis in HCXAI from explanation content to explanation motivation.

Our aim is to introduce a theoretical lens that we believe is currently missing from HCXAI discourse, and to invite the community to explore its implications.

Several research questions emerge from this perspective. 

First, for agentic AI systems that execute multi-step plans with cascading consequences, when is the right moment to offer explanations? Before execution, explanations may support planning but risk cognitive overload. During execution, they may enable intervention but fragment attention. After execution, they support accountability but cannot prevent harm. The timing of explanations may be as important as their content and the optimal timing likely depends on how users weigh the three utilities in a given moment.

Second, how do the three utilities interact in high-stakes contexts such as healthcare, criminal justice, or financial decision-making? In these domains, the tension between instrumental value (users need to act on explanations) and hedonic avoidance (explanations may reveal distressing risks) is particularly acute. Designing for one utility may compromise another.

Third, how do individual differences in information-seeking orientation shape explanation engagement?Understanding how the three utilities are weighted differently across user types could inform adaptive explanation strategies.

We invite the HCXAI community to integrate insights from the psychology of information seeking into future research. In particular, we propose that empirical work explore how instrumental, hedonic, and cognitive utilities interact in the context of agentic AI and how explanation design can better align with the ways people actually decide what they want to know to promote better design for HCXAI systems and facilitate the goals of fair human-AI teaming.

\section{Acknowledgments} This work is funded by 
NextGenerationEU programme under the funding schemes PNRR Partnership Extended PE00000013 - “FAIR - Future Artificial Intelligence Research” - Spoke 1, “Human-centered AI”, and TANGO project grant agreement no. 101120763



\bibliographystyle{ACM-Reference-Format}
\bibliography{sample-base}

\end{document}